\documentclass[letterpaper]{article} 
\usepackage{aaai25}  
\usepackage{times}  
\usepackage{helvet}  
\usepackage{courier}  
\usepackage[hyphens]{url}  
\usepackage{graphicx} 
\usepackage{natbib}  
\usepackage{caption} 
\usepackage{algorithm}
\usepackage{algorithmic}

\usepackage{amsmath}
\usepackage{amssymb}
\newtheorem{theorem}{Theorem}
\newtheorem{lemma}{Lemma}

\newtheorem{hypothesis}{Hypothesis}
\usepackage{multirow}
\usepackage{makecell}
\usepackage{adjustbox}
\usepackage{amsfonts}
\usepackage{enumitem}
\usepackage{varwidth}

\usepackage{newfloat}
\usepackage{listings}
\DeclareCaptionStyle{ruled}{labelfont=normalfont,labelsep=colon,strut=off} 
\floatstyle{ruled}
\newfloat{listing}{tb}{lst}{}
\floatname{listing}{Listing}
\title{\textbf{NBgE: a multi-physics-informed encoder leveraging bond graphs}}
\author{
    Alexis-Raja Brachet\textsuperscript{\rm 1, \rm 2}\thanks{Corresponding Author.}, Pierre-Yves Richard\textsuperscript{\rm 2}, Céline Hudelot\textsuperscript{\rm 1}
}
\affiliations{
\textsuperscript{\rm 1} MICS, CentraleSupélec, Université Paris-Saclay, France
\textsuperscript{\rm 2} CentraleSupélec, IETR UMR CNRS 6164, France
}

\begin{document}

\maketitle

\begin{abstract}
In the trend of hybrid Artificial Intelligence techniques, Physical-Informed Machine Learning has seen a growing interest. It operates mainly by imposing data, learning, or architecture bias with simulation data, Partial Differential Equations, or equivariance and invariance properties. While it has shown great success on tasks involving one physical domain, such as fluid dynamics, existing methods are not adapted to tasks with complex multi-physical and multi-domain phenomena. In addition, it is mainly formulated as an end-to-end learning scheme. To address these challenges, we propose to leverage Bond Graphs, a multi-physics modeling approach, together with Message Passing Graph Neural Networks. We propose a Neural Bond graph Encoder (NBgE) producing multi-physics-informed representations that can be fed into any task-specific model. It provides a unified way to integrate both data and architecture biases in deep learning. Our experiments on two challenging multi-domain physical systems - a Direct Current Motor and the Respiratory System - demonstrate the effectiveness of our approach on a multivariate time-series forecasting task.
\end{abstract}

%

\section{Introduction} \label{sect:introduction}

In recent years, deep learning has successfully processed data in various tasks. In the engineering field, recent attempts have shown the need to combine it with human knowledge, an approach named \textbf{Informed Machine Learning}. Our paper considers physical knowledge, positioning us in the Physical-Informed Machine Learning (PIML) field \cite{PIMLsurvey}.

PIML attempts can be divided into three main classes according to the type of inductive bias from knowledge: data, learning, and architecture bias  \cite{PIMLsurvey}. Data bias mainly consists of generating training data from simulations. A typical learning bias approach is Physical-Informed Neural Networks (PINNs) \cite{PINNraissi} leveraging Differential Equations in training objectives. Architecture bias incorporates physical properties such as equivariances and invariances into the neural models. These approaches have shown performances in fluid dynamics  \cite{PINNraissi} or material discovery  \cite{klipfel2023equivariant} in particular in the \textit{some data, some knowledge} paradigm  \cite{PIMLsurvey}. 

When multi-physical systems are involved, these models are still struggling. However, in the physical modeling field, some formalisms have been proposed to unify physical systems representations, among which the Bond Graph formalism \cite{bgpaynter1958generalizing} built on the shared notion of energetic power. It provides a domain-independent representation of any multi-physical systems. For instance, \cite{Rolle_Hernández_Richard_Buisson_Carrault_2005} model the whole cardiovascular system in a unique bond graph involving fluid mechanics, chemical reactions, electricity, and mechanical deformation. Nevertheless, to our knowledge, hybridization between Bond graphs and deep neural networks is under-studied. This paper is the first attempt to answer this limitation. 

We propose to answer the following Research Questions (RQ): \underline{\textbf{RQ1}}: can multi-physic bond graphs formalism be leveraged to inform ML models ?; \underline{\textbf{RQ2}}: can we develop a unique architecture for multi-physics instead of several physical domain-specific models ?; \underline{\textbf{RQ3}}: can we build a task-agnostic encoder using bond graph knowledge to inform any ML model? 

Following the current ML community trend, which consists of learning representations, we propose working on a physical-informed model-agnostic encoder. We introduce NBgE, a new architecture that combines bond graphs with Message Passing Graph Neural Networks (MPGNNs) and leverages data and architecture bias. 
We demonstrate its effectiveness on a multivariate time-series forecasting task on two challenging multi-physics systems: a Direct Current (DC) Motor and the Respiratory System (RS). This research scope excludes PINNs comparison, only informing Neural fields in end-to-end learning, hardly scalable for multi-physical systems (see \cite{multiphy-pinn}).


\begin{figure*}[htbp!]
  \includegraphics[width=0.95\textwidth]{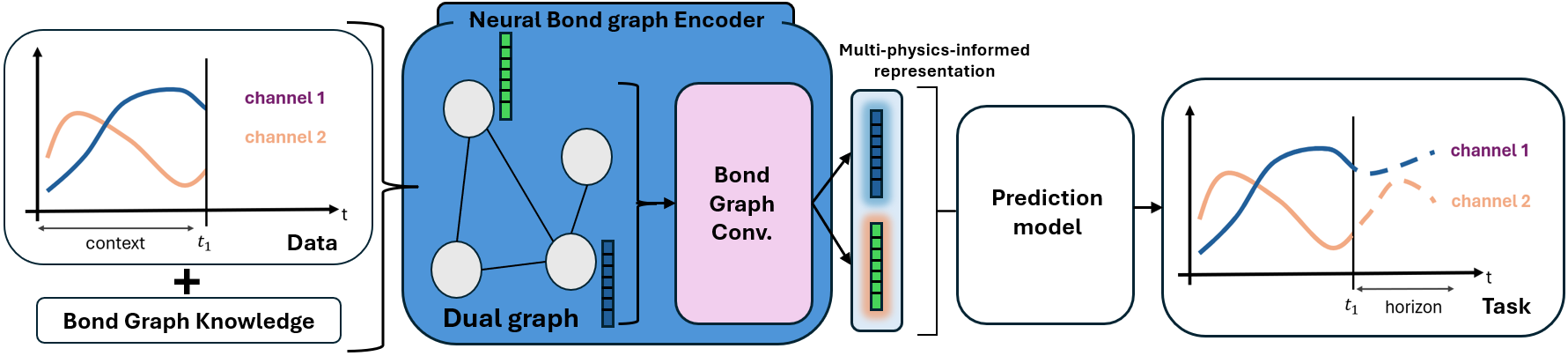}
  \caption{The NBgE pipeline. A dual graph is generated from the input data and the Bond Graph of the system, gathering the data and the physical knowledge. Then, NBgE encodes the input data with the Bond Graph Convolution defined on top of the dual graph. The multi-physics-informed representation can be taken as the input of any task-specific prediction model, such as here, in this paper, a time-series forecasting model.}
  \label{figure:pipeline}
\end{figure*}



We organize this paper as follows: in Section \ref{sect:theory} we recall the main principles of the bond graph formalism and MPGNNs. In Section \ref{nbge}, we develop the NBgE architecture with its three main components; (1) the Bond Matrix (Section \ref{bond matrix}), then (2) the dual graph generation process (Section \ref{graph generation}) and (3) the Bond Graph Convolution (Section \ref{bgc}), the core of the NBgE (Section \ref{building the nbge}), taking advantage of the relaxed bond graph knowledge (Section \ref{sect:knowledge-distillation}). We then conduct experiments in Section \ref{experimental analysis} to empirically validate the performances of the NBgE in the case of a DC motor (own simulated dataset) and the human RS (public dataset), two challenging multi-physical systems. Code, including the datasets, will be publicly available. We finish the paper with the related work in Section \ref{related-work} and a conclusion presenting our ongoing works. The overall pipeline is detailed in Figure \ref{figure:pipeline}.

\section{Theoretical preliminaries} \label{sect:theory}


\subsection{Preliminaries on Bond graphs} \label{intro_bg}

A bond graph $\mathcal{BG}(\mathcal{C},\mathcal{B})$ is a directed node-edge-attributed graph composed of a set of nodes $\mathcal{C}$ that represents the different components of the system and a set of bonds $\mathcal{B}\subseteq \mathcal{C}\times \mathcal{C}$ representing the power exchanges between the components.


Each bond $b$ has 4 attributes $(k, e, f, |)$: id, effort - flow variables, and causal stroke. $e$ and $f$ are physical signals. They are continuous and bounded with respect to time; thus, for any signal acquisition if time 
$T \in \mathbb{R}_+^*$, ${\int_{[0,T]} e(\tau)d\tau < +\infty}$ 
and ${\int_{[0,T]} f(\tau)d\tau < +\infty}$. Power is defined as ${P = e\times f}$. The direction of the bond sets the positive power exchange orientation. The causal stroke sets the direction of effort and flow signals between the two components (nodes) connected by the bond (edge). Given a bond connecting two components $\mathtt{C}_1$ and $\mathtt{C}_2$, if the causal stroke is at $\mathtt{C}_1$'s end, then $\mathtt{C}_1$ receives the effort from $\mathtt{C}_2$ and it gives back the flow and vice versa.

Each node $\mathtt{C}$ has 3 attributes $(c,k_c,\varphi)$: component type, id (depending on the component type), and the way it processes effort and flow variables of the edges connected to it.


A component $\mathtt{C} \in \mathcal{C}$ can be an element or a junction. There are 5 types of elements depending on the relation they impose between the effort and the flow \cite{Dauphin-TanguyGeneviève2000Lbg}. \textbf{Sources of effort} (SE) and \textbf{sources of flow} (SF) are source elements: effort and flow of the bonds attached to a source are independent. \textbf{Resistances} (R) are physical elements in which effort and the flow are related by the equation: $\varphi_R(e,f) = 0$; \textbf{inductances} (I) are components in which $\varphi_I(\int_t e(\tau) d\tau,f) = 0$; and \textbf{capacitances} (C) are components in which $\varphi_C(\int_t f(\tau) d\tau,e) = 0$.


Then, there are four types of junctions, distinguished by how power is distributed among the bonds connected at the same junction \cite{Dauphin-TanguyGeneviève2000Lbg}.
For any bonds $i$ and $j$ connected to a \textbf{0-junction} ,  ${\varphi_{0,1} (e_i,e_j) = e_i-e_j = 0}$ (unique characteristic effort). Additionally, flow is balanced between the $n$ bonds connected to it : ${\varphi_{0,2}(f_1,\dots,f_n) = \sum_i a_if_i = 0}$ with $a_i=1$ if the bond is directed towards the junction, $a_i = -1$ otherwise . \textbf{1-junctions} follow similar principles but with flows and efforts interchanged. Furthermore, \textbf{Transformers} (TF) are physical junctions in which efforts and flows of bonds $i < j$ connected to them are related such as ${\varphi_{TF}(e_i,e_j) = 0}$ and ${\varphi_{TF}(f_j,f_i) = 0}$. Similarly, \textbf{Gyrators} (GY) define the following relations ${\varphi_{GY}(e_i,f_j) = 0}$ and ${\varphi_{GY}(e_j,f_i) = 0}$.


One of the main advantages of this formalism is domain-agnostic property (RQ2), as power can be defined in any physical field. For instance, in electricity, effort is voltage, flow is current; in mechanics, $e$ is force and $f$ velocity; in fluid mechanics, $e$ is pressure and $f$ volumetric flow rate.

Another important aspect of the bond graph formalism is the causality, which sets how the equations are written to establish the state system.
In particular causality assignment rules \cite{Dauphin-TanguyGeneviève2000Lbg} guarantee the existence of a solution for the set of system equations, the main one being: \textit{exactly one bond, called the \textbf{strong bond}, imposes the effort (resp. flow) in 0-junction (resp. 1-junction)}.





\subsection{Preliminaries on MPGNNs} \label{mpnn}

MPGNNs belong to the class of deep learning algorithms acting on a graph $G(V,E)$, where $V$ are the nodes and $E$ the edges. They originate from the adaptation of Convolutional Neural Networks (CNNs) on grid graphs to a network acting on a graph with an arbitrary shape \cite{kipf2017semisupervised}. Their main principle consists of updating, within each layer $l$, the features $\textbf{h}^{l-1}(v)$ of each node $v$ as $\textbf{h}^{l}(v) = f_{\text{update}}^{\theta^l_u}(\textbf{h}^{l-1}(v),f_{\text{aggr}}^{\theta^l_a}(\{ \textbf{h}^{l-1}(u)|u\in N(v) \}))$, where $u\in N(v)$ are all the nodes such as $(u,v)$ is an edge. $f_{\text{update}}$ and $f_{\text{aggr}}$ define the MPGNNs architecture, i.e., how the information is processed across each node, $\theta^l_u$ and $\theta^l_a$ are parameters updated through the learning process. 

\section{NBgE : a multi-physics-informed encoder} \label{nbge}

We have seen bond graphs, which are graphical multi-physical representations operating on continuous physical signals stored in the bonds (edges). On the other hand, we have seen deep neural models operating on graphs where information is stored in the nodes. In this paper, we develop a method to translate the bond graphs into readable graphs for an MPGNN (RQ1 \& RQ2), where effort and flow variables are stored in nodes, and physical relations are stored in the edges. The MPGNN will produce a physical-informed representation of the input, which can be processed by any AI model (RQ3).
Before presenting this translation approach, we detail the assumptions about the physical system and associated bond graph.

\subsection{Working hypotheses} \label{sect:hypothesis}


\begin{hypothesis}[Existence of a connected Bond graph]\label{hyp:reduced-bg}
We assume that the studied phenomena are governed by physical laws that can be modeled by a connected bond graph with at least three components.
\end{hypothesis}


\begin{hypothesis}[1D bond graphs] \label{hyp:1Dsys}
Only 1D bond graphs are considered. In other words, a bond carries only one effort and flow variable;  physical elements (SE, SF, R, I, C) are 1-port elements; physical junctions (TF, GY) are 2-port junctions. The components (elements and junctions) define 1D constitutive relations.
\end{hypothesis}

\begin{hypothesis}[Linear physical relation] \label{hyp:lin-phy}
Only proportional element relations, i.e., first-order physical approximations, are considered. Under hypothesis \ref{hyp:1Dsys}, we have for $\mathcal{BG}(\mathcal{C,B})$, $\forall \mathtt{C} \in \mathcal{C} \, |\, c_{\mathtt{C}} \in \{ R,I,C,TF,GY \},  \exists \alpha_{\mathtt{C}} \in \mathbb{R^*}  \, |\, \varphi_{\mathtt{C}}(x,y) = x-\alpha_{\mathtt{C}} y = 0$.
\end{hypothesis}

 

\subsection{A matrix representation of a bond graph} \label{bond matrix}



In this section, we define a new representation of a bond graph on which the translation process will run. 

Given a bond graph $\mathcal{BG}(\mathcal{C}, \mathcal{B})$, let denote $M = card(\mathcal{B})$ the number of bonds in $\mathcal{BG}(\mathcal{C}, \mathcal{B})$, and $k_b\in [1,\dots,M]$ the id of $b\in \mathcal{B}$. For clarity, we will write $b$ instead of $k_b$ to identify matrices rows. We introduce the $M \times 9$ matrix $\mathcal{BM}$, called the Bond Matrix, representing the bond graph (DC motor's Bond Matrix is developed in Figure \ref{fig:bg2g}). The columns are divided into $2$ groups :

\begin{itemize}
    \item The first group is the \textit{physical element group}. If a bond $b \in \mathcal{B}$ is connected to a component $\mathtt{C} \in\mathcal{C}$ of type $c \in \{ SE,SF,R,I,C \}$, then $\mathcal{BM}[b,c] \neq 0$. If $c\in \{ SE,SF \}$, $\mathcal{BM}[b,c] = k_c$. If $c\in \{ R,I,C \}$, $\mathcal{BM}[b,c] = \alpha_{\mathtt{C}}$,
    \item The second group is the \textit{junction-causality group}. We employ triplets here as descriptors as follows: the algebraic id of the junction, the linear physical coefficient, and the variable imposed on the junction (i.e., causality).  If a bond $b$ is connected to a junction $\mathtt{C} \in\mathcal{C}$ of type $j \in \{ TF,GY,0,1 \}$, then $\mathcal{BM}[b,j] \neq 0$. 
    If $b$ is directed towards $j$, then $\mathcal{BM}[b,j][0] = +k_j > 0$, otherwise, it is equal to $-k_j<0$. If $j\in \{ TF,GY \}$, then $\mathcal{BM}[b,j][1] =  \alpha_{\mathtt{C}}$,  otherwise it is empty. If b's causal stroke is at $j$'s end, then $\mathcal{BM}[b,j][2] = e$, otherwise, it is $f$.
\end{itemize}
With this new representation of a bond graph, thanks to lemma \ref{lemma:junction} :

\begin{lemma}\label{lemma:junction}
Any bond is connected to a junction. 
\end{lemma}
\textit{Proof. } See Appendix.  $\Box$ 

We then have the following theorem :

\begin{theorem}[Expressiveness of the Bond Matrix]\label{expressivity}
Any bond graph is fully described by its bond matrix down to the $R,I,C$ elements' ids and effort-flow variables.
\end{theorem}
\textit{Proof. } See Appendix.  $\Box$

Effort and flow variables correspond to the input data. 
$R,I,C$ ids are not informative for us: we care about how they relate $e$ and $f$. This theorem allows us to work with the Bond Matrix for the dual graph generation.




\subsection{The Dual graph generation process from the Bond Matrix} \label{graph generation}


In this paper, we consider a multivariate time-series denoted $\mathbf{X} \in \mathbb{R}^{N \times D}$ with $N$ timestamps and $D$ aligned uni-modal channels. 
We assume $\mathbf{X}$ is generated by a (partially known) multi-physical system represented by a bond graph $\mathcal{BG}(\mathcal{C}, \mathcal{B})$ with input channel corresponding to an effort or a flow variable of a bond in $\mathcal{BG}(\mathcal{C}, \mathcal{B})$.

From now, we assume that $\mathcal{BG}(\mathcal{C}, \mathcal{B})$ is complete
\footnote{\textbf{Definition 1 (Completeness) }.
A bond graph $\mathcal{BG}(\mathcal{C}, \mathcal{B})$ with $M$ bonds is complete with respect to $\mathbf{X}$ if and only if there exists an injective function $f_{\mathbf{X},\mathcal{BG}}$ such as :\begin{eqnarray}\forall d \in \{ 1,\dots ,D \}, \exists! (b,var) \in \mathcal{B}\times \{e;f\} \space | \space f_{\mathbf{X},\mathcal{BG}}(d) = var_{b}\end{eqnarray}}
with respect to $\mathbf{X}$ and denote $\mathcal{BM}$ the corresponding Bond Matrix. The goal here is to generate a new graph $G(V,E)$ representing the data and the multi-physical bond graph knowledge to work on the latent representation of the input channels by performing physical-informed message passing. 



Our methodology involves: (a) a translation into the frequency domain through the Discrete Fourier Transform (DFT) to manage the temporal integrations involved by the physical relations (see Section \ref{intro_bg}) as frequency multiplications, naturally handled by neural networks \cite{DBLP:FNO}; (b) a 7-step algorithm taking as input the bond matrix $\mathcal{BM}$, $\mathbf{\hat{X}}$ (DFT of $\mathbf{X}$), and $f_{\mathbf{X},\mathcal{BG}}$ and outputs a graph $G$ which translates the bond graph knowledge in a dual fashion. Note that representations in the frequency domain are sparse and add interpretability \cite{DBLP:FNO,DBLP:FEDformer,yi2023survey}

\begin{figure*}[t]
\centering
\includegraphics[width=0.95\textwidth]{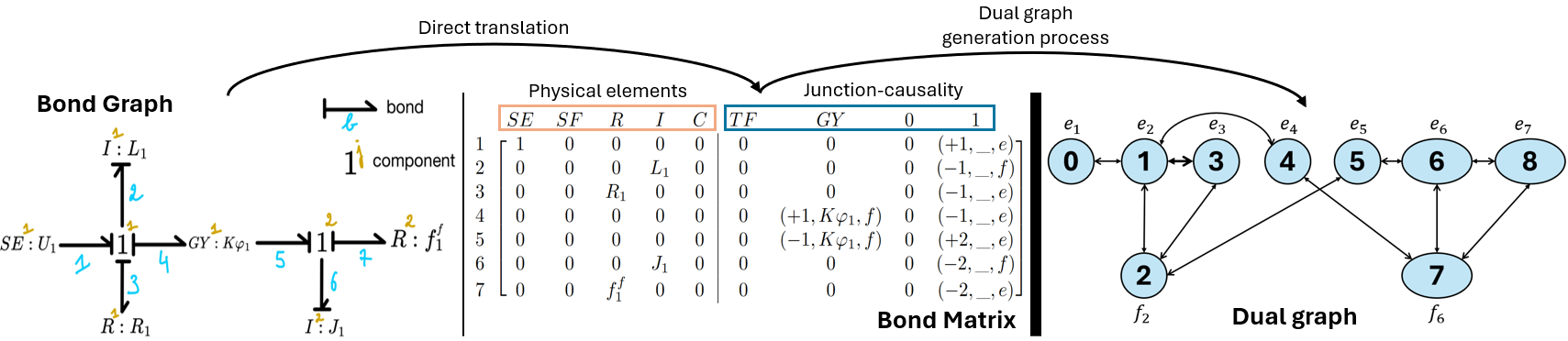} 
\caption{From the physical system to the dual graph. The case of the DC Motor with 7 bonds is illustrated here. From the bond graph of the system, the Bond Matrix is developed. Then, through the 7-step process, the dual graph is generated. The nodes represent the effort and flow variables, the edges represent the way the power is distributed. Once the dual graph is initialized, the Bond Graph Convolution updates the node features, being the input of any task-specific model.}
\label{fig:bg2g}
\end{figure*}

\paragraph{a. Discrete frequency matrix operators}


We introduce the integration and derivation discrete frequency matrix operators, with $N$ the number of timestamps and $f_s \in \mathbb{R_+^*}$ the sampling frequency. By using the trapezoidal rule for integration, the Z-transform, and linearization, we obtain $\mathcal{I}_N = \textrm{diag}(0,\frac{1}{2\pi f_s}\frac{N}{i k})_{k \in 1,\dots, N-1}$ for the integration and  $\mathcal{D}_N = \textrm{diag}(2\pi f_s\frac{ik}{N})_{k \in 0,\dots, N-1}$ for the derivation. The important point here is that physical relations $\varphi$ defined in the bond graphs with temporal integration and derivation become matrix operations with $\mathcal{I}_N$ and $\mathcal{D}_N$. 

\noindent For instance,  with $f(t) = \alpha_I^{-1}\int_0^t e(\tau)d\tau$, we get ${\hat{\textbf{f}}_N = \alpha_I^{-1}\mathcal{I}_N \hat{\textbf{e}}_N}$ for an I-element with $\hat{\textbf{f}}_N$ and $\hat{\textbf{e}}_N$ the DFT vectors of $f$ and $e$. These physical relations in the discrete frequency domain will define the edge features of the dual graph $G$: $\alpha_I^{-1}\mathcal{I}_N$ would be the feature of the edge going from node $e$ to $f$ for an I-element. 

\paragraph{b. Dual graph generation process}

We develop a 7-step procedure to build the graph $G$ from $\mathcal{BM}$ (The DC Motor dual graph is shown in Figure \ref{fig:bg2g}). In the bond graph formalism, information is stored in the edges, and the nodes express how the power is exchanged from one bond to another. From the bond matrix, we translate the bond graph into a new graph, readable for an MPGNN, where information is on the nodes with the edges expressing information exchange. First, (1) we create one node per variable. Then, (2) we add directed edges (direction is set by causality) based on 1-port elements (R, I, C) relations - link between effort and flow of the same bond -, then (3) edges for 2-port ones (TF, GY) - link between effort and flow of two different bonds attached to the same TF-GY -, and (4) edges for n-port ones (0-1 junctions) - link between multiple bonds effort and flow attached to the same 0-1 junction -. We then (5) merge nodes representing the same variable according to the strong bond imposing effort-flow on 0-1 junctions. We then (6) relax causality to add freedom to the MPGNN, allowing messages to be exchanged in both directions. As all physical signals are not necessarily measured ($f_{\mathbf{X},\mathcal{BG}}$ is injective), the final step (7) completes the missing node features of $G$. 

The directed node-labeled and node-edge-attributed graph $G(V,E,\mathcal{L}_V,\mathbf{H}_V,\mathcal{F}_E)$ is completely defined (RQ2). We can perform message passing over it thanks to the Bond Graph Convolution (BGC), a physical-informed MPGNN.






\subsection{The Bond Graph Convolution} \label{bgc}


\begin{algorithm}[tb]
\caption{The Bond Graph Convolution at a layer l}\label{alg:BGC}
\begin{flushleft}
\textbf{Input}: $G$ (dual graph), $\mathbf{H}_V^{l-1}$ (node features at layer $l-1$)\\
\textbf{Output}: $\Tilde{\mathbf{H}}_V^{l}$ \text{(node features at layer l)}
\end{flushleft}
\begin{algorithmic}[1] 
\STATE $\hat{\mathbf{H}}_{Vs}^{l}\gets \mathbf{0}_{|V|\times d_l}$
\STATE $\hat{\mathbf{H}}_{Vs}^{l-1} \gets \texttt{Sampling}_l(DFT(\mathbf{H}_V^{l-1}))$
\FOR{i in V}

\STATE  $ \hat{\mathbf{H}}_{Vs,i}^{l} \gets \frac{1}{\texttt{card}(\mathcal{N}(i)\cup \{ i \})} (\sum_{j \in \mathcal{N}(i)} \hat{\mathbf{H}}_{Vs,j}^{l-1} \Phi^l_{s,(j,i)} +\hat{\mathbf{H}}_{Vs,i}^{l-1}) $

\ENDFOR
\STATE $\hat{\mathbf{H}}_V^{l} \gets \texttt{Padding}_l(\hat{\mathbf{H}}_V^{l-1},\hat{\mathbf{H}}_{Vs}^{l})$
\STATE $\Tilde{\mathbf{H}}_V^{l} \gets DFT^{-1}(\hat{\mathbf{H}}_V^{l})$
\end{algorithmic}
\end{algorithm}

Instead of tuning a global message passing over a graph as in \cite{kipf2017semisupervised,sage,gat}, we learn a specific message passing over each physical-informed edge. We use the same notations as in Section \ref{mpnn}: $f_{\text{update}}$ is the mean function, and $f_{\text{aggr}}$ is the same as in \cite{kipf2017semisupervised} but here,
the message passing depends on the edges as the matrices processing the messages are the edge features $\{ \Phi_{(u,v)}\}_{(u,v)\in E}$ updated through the learning process. The BGC pseudo-code is developed in Algorithm \ref{alg:BGC}.

We choose to model the linear bond graph relations in the discrete frequency domain as an unbiased 1-layer Feed forward network (FFN) $\Phi$ initialized with the known physical relations. In the previous example, if $b \in \mathcal{B}$ is connected to an I-element, we would have $\Phi_{(e_b,f_b)}^{\text{init}} = \alpha_I^{-1}\mathcal{I}_{N}$, where $e_b$ and $f_b$ are the nodes of bond $b$'s effort and flow and $N$ the number of timestamps.
The $DFT$ computations are done within the BGC. We perform \texttt{Sampling} (defined in \cite{DBLP:FEDformer}) and \texttt{Padding} operations to make the most of the sparse property of the Fourier Transform. The \texttt{Padding} consists of updating the selected modes from the \texttt{Sampling} on the input features $\hat{\mathbf{H}}_V^{l-1}$ (instead of padding with zeros). The number of sampled modes can vary from one layer to another. 
In Algorithm \ref{alg:BGC}, the hat identifies the frequency domain, and the subscript $s$ stands for \textit{sampled}.

During the learning process, $\{ \Phi_{(u,v)}^l\}_{(u,v)\in E}$ are updated: non-linear physical phenomena can be learned as non-diagonal weights appear.

\subsection{Building the Neural Bond graph Encoder} \label{building the nbge}

The data goes first in a scaler layer of dimension $d_{0}$ \cite{DBLP:FNO} (layer $0$).
From the BGC, we build a Bond Graph Layer (BGL) :

\begin{eqnarray}
&\Tilde{\mathbf{H}}_V^{l} = BGC^l(\mathbf{H}_V^{l-1}) \\
&\mathbf{H}_V^{l} = \sigma((\alpha_{BGC} \cdot \Tilde{\mathbf{H}}_V^{l}+\alpha_{\text{skip}}\cdot\mathbf{H}_V^{l-1})W^l + B^l) \label{eq4}
\end{eqnarray}
with $\sigma$ the ReLU function, $W^l \in \mathbb{R}^{d_{0} \times d_{0}}$ and $B^l \in \mathbb{R}^{d_{0}}$. $\alpha_{BGC}$ and $\alpha_{\text{skip}}$ are weighting coefficients verifying $\alpha_{BGC} + \alpha_{\text{skip}} = 1$.

Then we perform Cross Attention \cite{MTGODE} to compute the output of the Neural Bond graph Encoder (NBgE): 
\begin{eqnarray}
\mathbf{H}_V^{out} = \sum_{l=1}^L (\mathbf{H}_V^{l}W_a^l+B_a^l)
\end{eqnarray}

with $W_a^l \in \mathbb{R}^{d_0\times d_0}$ and $B_a^l \in \mathbb{R}^{d_0}$. $\mathbf{H}_V^{out}$, a multi-physics-informed representation of $\mathbf{X}$, can be fed into any task-specific model (RQ3).

\subsection{Bond graph knowledge distillation in NBgE} \label{sect:knowledge-distillation}

We propose here a small analysis, illustrated with the DC Motor case, to show how the knowledge is distilled within NBgE. For any physical variable, the BGC update is:

\begin{eqnarray}
    h^{\text{updated}}_{\text{phy var}} = \underbrace{f_{\text{update}}( h_{\text{phy var}},f_{\text{aggr}}}_{\text{MPGNN}}(\underbrace{m_{\text{causal}}}_{\text{causal bg eq.}}, \underbrace{m_{\text{rev causal}}}_{\text{extended bg eq.}}))
\end{eqnarray}

\noindent with $m_{\text{causal}}$ and $m_{\text{rev causal}}$ the direct and reverted causal multi-physics-informed messages, $f_{\text{aggr}}$ the MPGNN aggregation (message summation) and $f_{\text{update}}$ the MPGNN update (average including skip connexion $h_{\text{phy var}}$) functions.

In the DC Motor bond graph equations, the effort of the strong bond 2 appears in two equations: $e_2 = e_1 - e_3 - e_4 \text{ (1-junction) }$ and $f_2 = \frac{1}{L_1} \int_{[0;T]} e_2(\tau)d\tau \text{ (inductance) }$. Following causality, $e_2$ should only be computed through the $1$-junction equation - bond $2$ receives the effort and gives back the flow -. 
Following the BGC update for $h_{e_2}$:
\begin{eqnarray}
        h^{\text{updated}}_{e_2} = [ h_{e_1}\Phi_{(e_1,e_2)} + h_{e_3}\Phi_{(e_3,e_2)} + 
        h_{e_4}\Phi_{(e_4,e_2)} \nonumber \\ + h_{f_2}\Phi_{(f_2,e_2)} + h_{e_2} ] /5
\end{eqnarray}

\noindent We can identify the causal bond graph message $m_{\text{causal}} = h_{e_1}\Phi_{(e_1,e_2)} + h_{e_3}\Phi_{(e_3,e_2)}  + h_{e_4}\Phi_{(e_4,e_2)}$ (1-junction) and the additional reverted causal message $m_{\text{rev causal}} = h_{f_2}\Phi_{(f_2,e_2)}$ (inductance).
Similarly, for bond $4$ (not a strong bond), $e_4$ appears in two equations : $e_2 = e_1 - e_3 - e_4$ (1-junction) and $e_4 = K\varphi_1 f_6$ (gyrator). According to causality, $e_4$ should only be computed by the gyrator equation. 
As the dual graph is built around the strong bonds, causality reversal creates incomplete relations as :
\begin{eqnarray}
    h^{\text{updated}}_{e_4} =&  \frac{h_{f_6}\Phi_{(f_6,e_4)}  + h_{e_2}\Phi_{(e_2,e_4)} + h_{e_4}}{3}
\end{eqnarray}

\noindent
Here, $m_{\text{rev causal}} = h_{e_2}\Phi_{(e_2,e_4)}$. $e_4$ only receives a signal from $e_2$ from the $1$-junction equation, where it should have received messages from the other efforts (according to causality reversal: ${e_4 = e_1 - e_2 - e_3}$).
With multiple BGL in series, $e_4$ will receive the other efforts information through $e_2$, so the $1$-junction equation shape is recovered. 

Adding $m_{\text{rev causal}}$, $f_{\text{aggr}}$, and $f_{\text{update}}$ aims to relax the bond graph constraints, hence giving more expressivity to NBgE as it takes advantage of all the bond graph equations and the MPGNN aggregation and update rules, which are blended through the BGC and the BGL in series (RQ1).




\section{Experimental study} \label{experimental analysis}

\begin{table*}[t]
  \centering
    \begin{tabular}{|c|c|cc|cc|cc|cc|}
    \hline
         \multicolumn{2}{|c|}{Study case}  &  \multicolumn{2}{|c|}{Linear}  & \multicolumn{2}{|c|}{MLP}  & \multicolumn{2}{|c|}{GraphSAGE}  & \multicolumn{2}{|c|}{Transformer} \\ \hline \hline

         \multicolumn{2}{|c|}{DC motor}  & w/o & w/ & w/o & w/ & w/o & w/ & w/o & w/  
         \\ \hline
        \multirow{4}{*}{\makecell{100 \\ - \\ 500}} & MAE  & 3.91 $\pm$ 0.00 & \textbf{1.64 $\pm$ 0.03} & 1.53 $\pm$ 0.04 & \textbf{1.37 $\pm$ 0.03} & 1.48 $\pm$ 0.01 & \textbf{1.32 $\pm$ 0.02} & \underline{\textbf{1.31 $\pm$ 0.04}} & 1.35 $\pm$ 0.05 \\
         & MSE & 52.7 $\pm$ 0.1  & \textbf{13.1 $\pm$ 0.4} & 14.7 $\pm$ 0.5 & \textbf{11.3 $\pm$ 0.4} & 16.0 $\pm$ 0.5 & \textbf{10.2 $\pm$ 0.5} & \underline{\textbf{9.70 $\pm$ 0.49}} & 10.2 $\pm$ 0.9 \\ 
         & SDTW & 2034 $\pm$ 7 & \textbf{455 $\pm$ 16} & 598 $\pm$ 28 & \textbf{369 $\pm$ 21} & 626 $\pm$ 24 & \textbf{356 $\pm$ 15} & \underline{\textbf{279 $\pm$ 26}}  & 296 $\pm$ 53 \\
         & \# param & 0.051 & 1.1 & 2.2 & 4.1 & 4.8 & 6.1 & 26.3 & 28.6 \\ \hline
         
         \multirow{4}{*}{\makecell{300 \\ - \\ 300}} &  MAE & 3.06 $\pm$ 0.01 & \textbf{1.26 $\pm$ 0.02} & \textbf{1.26 $\pm$ 0.02} & 1.27 $\pm$ 0.03 & \textbf{1.17 $\pm$ 0.02} & 1.23 $\pm$ 0.04 & \underline{\textbf{0.96 $\pm$ 0.07}} & 0.98 $\pm$ 0.02 \\
         & MSE & 33.2 $\pm$ 0.1 & \textbf{9.35 $\pm$ 0.64} & \textbf{7.44 $\pm$ 0.26} & 7.66 $\pm$ 0.30 & \textbf{6.99 $\pm$ 0.18} & 7.30 $\pm$ 0.22 & 5.87 $\pm$ 0.65 & \underline{\textbf{5.58 $\pm$ 0.44}} \\ 
         & SDTW & 673 $\pm$ 7 & \textbf{237 $\pm$ 18} & \textbf{204 $\pm$ 11} & 207 $\pm$ 9 & \textbf{180 $\pm$ 4} & 189 $\pm$ 7 & \underline{\textbf{103 $\pm$ 14}} & 105 $\pm$ 9 \\
         & \# param & 0.09 & 2.0 & 2.3 & 4.2 & 4.8 & 7.5 & 26.3 & 28.2 \\ \hline
         
         \multirow{4}{*}{\makecell{500 \\ - \\ 100}} & MAE & 1.81 $\pm$ 0.01 & \textbf{0.89 $\pm$ 0.02} & \textbf{0.60 $\pm$ 0.03} & 0.66 $\pm$ 0.03 & \underline{\textbf{0.52 $\pm$ 0.01}} & 0.64 $\pm$ 0.03 & 0.71 $\pm$ 0.06 & \textbf{0.71 $\pm$ 0.04} \\
         & MSE & 11.7 $\pm$ 0.1 & \textbf{5.53 $\pm$ 0.37} & \textbf{2.86 $\pm$ 0.30} & 3.90 $\pm$ 0.43 & \underline{\textbf{2.58 $\pm$ 0.08}} & 3.70 $\pm$ 0.34 & 3.75 $\pm$ 0.50 & \textbf{3.51 $\pm$ 0.41} \\ 
         & SDTW & 146 $\pm$ 2 & \textbf{63.1 $\pm$ 5.2} & \textbf{32.3 $\pm$ 5.4} & 41.8 $\pm$ 6.5 & \textbf{31.7 $\pm$ 1.2} & 39.5 $\pm$ 3.7 & 24.4 $\pm$ 3.3 & \underline{\textbf{22.7 $\pm$ 3.0}} \\
         & \# param & 0.051 & 3.0 & 2.4 & 5.0 & 4.8 & 7.6 & 26.3 & 27.8 \\ \hline \hline

         \multicolumn{2}{|c|}{\makecell{RS}}  & w/o & w/ & w/o & w/ & w/o & w/ & w/o & w/  \\ \hline
         \multirow{4}{*}{\makecell{100 \\ - \\ 500}} & MAE  & 12.5 $\pm$ 0.0  & \textbf{11.7 $\pm$ 0.0} & 13.1 $\pm$ 0.1 & \underline{\textbf{11.6 $\pm$ 0.0}} & 12.1 $\pm$ 0.10 & \textbf{11.8 $\pm$ 0.1} & \textbf{11.7 $\pm$ 0.0} & 11.9 $\pm$ 0.2 \\ 
          & MSE & 3.27 $\pm$ 0.00 & \textbf{2.98 $\pm$ 0.03} & 3.36 $\pm$ 0.05 & \underline{\textbf{2.86 $\pm$ 0.02}} & 3.13 $\pm$ 0.06 & \textbf{2.92 $\pm$ 0.05} & \textbf{3.00 $\pm$ 0.06} & 3.08 $\pm$ 0.09 \\ 
          & SDTW & 64.3 $\pm$ 0.1 & \textbf{54.0 $\pm$ 1.3} & 62.1 $\pm$ 2.3 & \textbf{59.0 $\pm$ 0.6} & 61.4 $\pm$ 1.8 & \textbf{56.2 $\pm$ 1.7} & 55.0 $\pm$ 3.9 & \underline{\textbf{52.7 $\pm$ 3.0}} \\
         & \# param & 0.051 &  10.7 & 1.4 & 11.9 & 4.2  & 18.6  & 8.4 & 28.1 \\  \hline
         
         \multirow{4}{*}{\makecell{300 \\ - \\ 300}} &  MAE & 11.7 $\pm$ 0.01 & \underline{\textbf{9.74 $\pm$ 0.05}} & 11.0 $\pm$ 0.18 & \textbf{9.87 $\pm$ 0.12} & 10.3 $\pm$ 0.06 & \textbf{10.2 $\pm$ 0.1} & 10.7 $\pm$ 0.3 & \textbf{10.6 $\pm$ 0.2} \\
         & MSE & 2.95 $\pm$ 0.01 & \underline{\textbf{2.32 $\pm$ 0.03}} &  2.52 $\pm$ 0.07 & \textbf{2.37 $\pm$ 0.06} & \textbf{2.32 $\pm$ 0.05} & 2.38 $\pm$ 0.06 & 2.80 $\pm$ 0.16 & \textbf{2.62 $\pm$ 0.09} \\ 
         & SDTW & 28.3 $\pm$ 0.1 & \textbf{21.8 $\pm$ 0.2} & 26.4 $\pm$ 0.6 & \underline{\textbf{20.9 $\pm$ 0.7}} & 24.3 $\pm$ 0.8 & \textbf{22.0 $\pm$ 0.8} &  23.3 $\pm$ 2.5 & \textbf{23.1 $\pm$ 1.6} \\
         & \# param & 0.09 & 9.7 & 1.4 & 11.0 & 4.2 & 18.1 & 8.4 & 70.8\\  \hline
         
         \multirow{4}{*}{\makecell{500 \\ - \\ 100}} & MAE & 10.3 $\pm$ 0.0 & \textbf{8.84 $\pm$ 0.16} & \textbf{8.70 $\pm$ 0.18} & 8.72 $\pm$ 0.14 & 8.75 $\pm$ 0.17 & \textbf{8.75 $\pm$ 0.12} & 8.58 $\pm$ 0.29 & \underline{\textbf{8.37 $\pm$ 0.13}} \\
         & MSE &  2.47 $\pm$ 0.01 & \textbf{1.76 $\pm$ 0.06} & \underline{\textbf{1.67 $\pm$ 0.05}} & 1.72 $\pm$ 0.06 & \textbf{1.70 $\pm$ 0.06} & 1.71 $\pm$ 0.04 & 1.79 $\pm$ 0.12 & \textbf{1.73 $\pm$ 0.09} \\ 
         & SDTW & 8.78 $\pm$ 0.09 & \textbf{6.77 $\pm$ 0.15} & \textbf{6.53 $\pm$ 0.40} & 6.58 $\pm$ 0.27 & \textbf{5.95 $\pm$ 0.23} & 6.57 $\pm$ 0.23 & \underline{\textbf{5.50 $\pm$ 0.67}} & 5.54 $\pm$ 0.26\\
         & \# param & 0.051 & 26.8 & 1.4 & 28.2 & 4.2 & 14.8 & 8.4 & 18.0 \\ \hline
         
    \end{tabular}
    \caption{Non-informed models (w/o) performances compared to the multi-physical-informed ones with NBgE (w/). For each metric, the lower, the better. The best scores between the informed and non-informed models are in \textbf{bold}, and the overall best scores are \underline{underlined}. (parameters are $\times 10^{6}$, SDTW are $\times 10$, RS's MAE and MSE score are $\times 10^{-2}$). Metrics were computed over the $10$ best out of $20$ consecutive runs.}
    \label{tab:results}
\end{table*}


\subsection{Datasets}

We test NBgE on two challenging physical systems: our own simulated dataset of a Direct Current Motor (fully-known physics) and the Respiratory System (RS) taken from \cite{lungs_data} (partially-known physics). 

\subsubsection{DC motor}
It has two domains: the electrical and mechanical ones. In the electrical part, a voltage source $U_1$ powers up a resistance $R_1 = 5 $ $\Omega$ and an inductance $L_1 = 0.1$H in series and an electro-mechanical converter ${K\varphi_1 = 0.1 \text{ V.s.rad}^{-1}}$ attached to a shaft $J_1 = 0.01 \text{ kg.m}^2$ subjected to fluid fiction $f^f_1 = 0.001 \text{ N.m.s.rad}^{-1}$ (the bond graph and bond matrix are developed in Figure \ref{fig:bg2g}). The DC motor model and simulation are performed on MATLAB SIMULINK \cite{MATLAB}. We set the voltage source to generate the signals as a square signal going from $0$ to $2$ V with a variable frequency and pulse sweep. We add some controlled noise: every $10$ seconds, $U_1$ is multiplied by a constant uniformly sampled between $0.8$ and $1.2$. Only the voltage current and the shaft rotation speed are accessible. The problem is ill-posed as the DC motor equations require the voltage source to be solved numerically.


\subsubsection{Respiratory System} \label{sect:resp-sys}

It has two domains: the fluid and the mechanical ones. The former corresponds to the air and the latter to the trachea distension, the lung movement, the muscle action, and the chest and abdomen movement. \cite{lungs_data} has recorded the data across $80$ people to develop automatic methods to monitor respiratory diseases. A complete bond graph is needed to employ our methodology. Our starting point is the bond graph of \cite{bg_lungs}, where the trachea and the lungs are modeled. The chest-abdomen part is missing. We conducted a power transmission reasoning to complete it: the lung swelling - modeled as a balloon - induces the chest and abdomen displacement thanks to the diaphragm. The accessible channels are the air pressure and flow in the trachea and the chest and abdomen displacement. We normalize the data of each channel between $0$ and $1$ to set the unknown bond graph parameter values to $1$.

\subsection{Experimental Setup}

NBgE performances are evaluated through a forecasting task. Let ${\mathbf{X} = \{ X^t_1, \dots , X^t_D \}_{t=1}^N}$ denote the historical data of $N$ timestamps. The forecasting task is to infer $K$ future timestamps based on the $N$ historical ones ${\mathbf{\Tilde{X}} = \{ \Tilde{X}^t_1, \dots , \Tilde{X}^t_D \}_{t=N+1}^{N+K}}$. 
On both datasets, to build up the samples for the task, we set a window of $600$ points per channel, and we extract them (without overlaps) from randomly chosen recordings to get a total of around $500$ samples for the task.
We test the models on three scenarios with decreasing difficulty : (N=100;K=500), (N=300;K=300), and (N=500;K=100). The scenarios share the same samples to allow direct comparison. To assess the NBgE model-agnostic property, we choose $4$ models being the basis of the majority of AI models: Linear \cite{linear_model}, Multilayer Perceptron (MLP), GraphSAGE \cite{sage} and Transformer \cite{DBLP:journals/corr/VaswaniSPUJGKP17} adapted to time-series in \cite{DBLP:FEDformer}. 
We omit PINNs in our comparison, which are state-of-the-art in the physical-informed field when PDEs are available. They are out of our research scope since they 
are not encoder-informed approaches (end-to-end learning scheme). NBgE and PINNs are not concurrent models and could be combined (see conclusion).

We take one configuration performing well for each model on each physical system and compare their performance across the setups with and without the NBgE. The models and learning hyperparameters are tuned with Tune and HyperOpt help \cite{bergstra2013hyperopt,liaw2018tune}.

In each case, we take $80 \%$ of the samples for training ($70 \%$ train  - $10 \%$ validation) and $20 \%$ for testing. We choose the Huber Loss ($\delta = 0.1$) to train all the models based on \cite{huber_loss}. Our metrics comparison are the Mean Average Error (MAE), the Mean Square Error (MSE), and the normalized Soft Dynamic Time Warping (SDTW) ($\gamma=0.1$) \cite{cuturi2018softdtw}. The first two are basic metrics for time-series forecasting tasks; the latter controls shape similarity. 

\subsection{Overall Comparisons}

\begin{figure*}[h!]
\centering
  \includegraphics[width=0.9\textwidth]{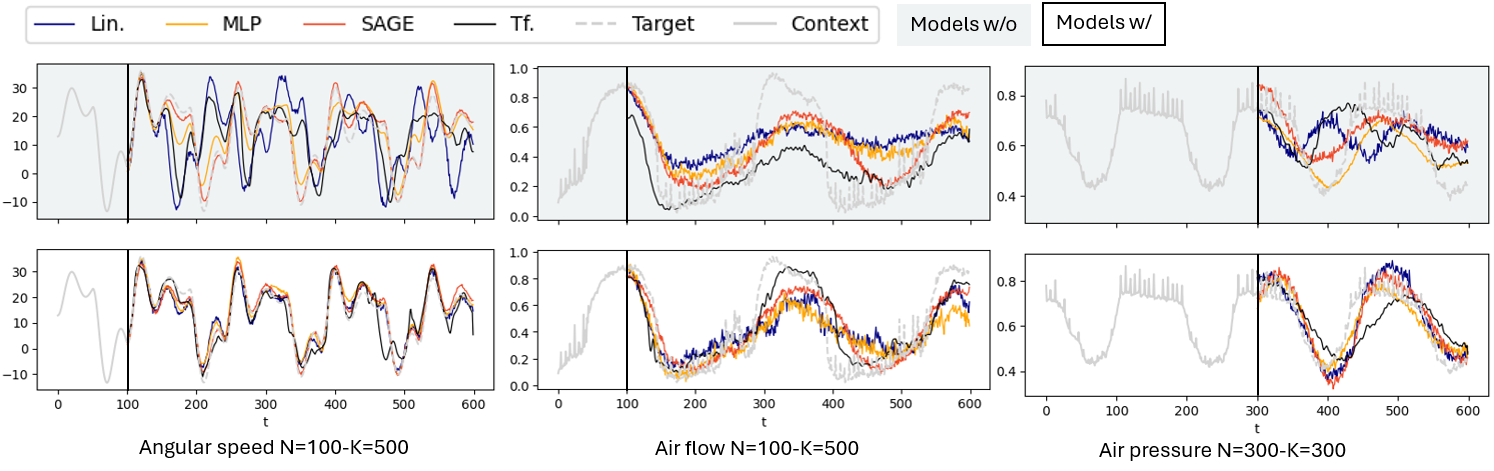}
  \caption{Predictions on three windows covering the studied physical systems: the DC Motor's angular speed and the RS's two air channels. The input is in gray, and the target is in dashed gray. NBgE-informed model predictions, on the white background, are shown against non-informed ones, on the gray background. 
  }
  \label{fig:courbes}
\end{figure*}

We can observe, from the results in Table \ref{tab:results}, 
NBgE always improves the Linear results. As the task difficulty increases, a performance tendency of NBgE-informed over non-informed models is observed for MLP and GraphSAGE, going from failure to success. The crossing line in the DC-Motor case is around (N=300;K=300) setup and around (N=500;K=100) for the RS, both where metric intervals start to overlap. Bond graph constraints are relaxed on purpose, giving NBgE the freedom to get hidden properties. It seems misleading in the DC Motor with simple dynamics where non-informed models have enough data. In the RS, knowledge from the bond graph is approximative, and the task is harder: relaxation appears effective. NBgE seems relevant in such difficult setups, motivating its use (RQ1). This could correspond to the \textit{some physics, some data} setup where PIML shines \cite{PIMLsurvey}. Despite the limited knowledge of the RS, NBgE generates more informative representations of the data in such a setup: NBgE-informed Linear, MLP, and GraphSAGE get better results than the non-informed Transformer, which again would support the encoder's benefits (RQ1). Furthermore, Transformers follow opposite trends. Due to their larger capacity, while knowledge is less needed in (N=100; K=500), non-informed Transformers overfit as the task gets easier in the (N=300; K=300)-(N=500; K=100) setups where NBgE helps them to generalize with competitive scores and lower standard deviations. Across each model (except Linear, which is quite basic), NBgE tends to stabilize the predictions when it gets similar or better results over non-informed models.

\section{Related work} \label{related-work}


\paragraph{Physical-Informed Neural Networks} 
PINNs architectures \cite{PINNraissi} take as input the space-time location and output the state of the studied system at the input location. The residuals of the equations governing the data are added to the loss of the network during training: the more physical domains, the more equations, the more hyperparameters to tune. There isn't any unified framework for multiphysics yet \cite{PIMLsurvey}. For instance, in \cite{multiphy-pinn}, heat transfer is studied between the air and a wall. To address the multi-physical domain, they define two separate PINNs - one for the fluid and one for the solid domain - coupled by a global loss trained for the solid and fluid domain. 

\paragraph{Physical-Informed Graph Neural Networks} Graph Neural Networks (GNNs) are widely used for physical applications as they are equivariant to permutation by design. However, one would like the model to have additional equivariances transcribing the system's physical properties. In \cite{satorras2022en,klipfel2023equivariant,zhang2023protein,regio_elasticity}, they design a specific message passing algorithm on the molecular graph, invariant and equivariant to targeted transformations. In \cite{morris2024orbitequivariant}, they define a new class of GNNs to relax the equivariance to permutation.


\paragraph{Bond graphs and ML} Several papers have worked on combining Bond graphs and ML models. The bond graph formalism models each part of the studied system, which is interesting in a Fault Detection and Isolation (FDI) framework. Indeed, with bond graphs, we can identify and isolate the causes of the system's faulty behavior, which is crucial for maintenance purposes. In \cite{BGCNN}, they combine the residuals generated by the bond graph \cite{ARRoriginalArt} and a CNN to classify the fault and compute the contribution of each part of the system to it. In \cite{BGbayesian}, they generate a Bayesian network from the bond graph and compute the probability of being dysfunctional over each bond graph component. In the case of complex systems, bond graphs can suffer from numerical issues such as algebraic loops \cite{Dauphin-TanguyGeneviève2000Lbg}. AI models can help bond graphs to simplify the equations and solve the system's governing equations \cite{dim-reduc-bg}.
The associations between bond graphs to help AI models are restricted to the FDI task. No attempts have been made to develop a hybrid Bond graph and AI formalism method in an informed framework for general tasks like forecasting.

\section{Conclusion}


We propose the first contribution leveraging the bond graph formalism through MPGNNs (RQ1). We introduce NBgE, a unified framework (RQ2) that can inform any task-specific model (RQ3) and tackle any bond graphs multi-physics systems (under the working hypotheses). We show the benefits of our approach on the time-series forecasting task on two datasets:  a simulated (DC motor) and a real-measured one (RS \cite{lungs_data}), especially when the knowledge is approximative. A clear trend confirms the benefits of employing NBgE in the \textit{some physics, some data} setup \cite{PIMLsurvey} (RQ1). 

This work is a first attempt, with opportunities to get more potent results.
Potential applications are possible, such as training NBgE in an unsupervised way as BERT \cite{devlin2019bert} and employing it in a downstream task or extending NBgE to fluid dynamics \cite{bg_fluid}. Finally, combining NBgE and PINNs by adding NBgE as a processing block in a PINN-learning scheme would leverage all the PIML inductive biases in an end-to-end pipeline, with NBgE helping PINNs on multi-physical systems. All of these points are considered as future work for us.

\bibliography{aaai25}

\end{document}